%% file: acl_latex.tex
\pdfoutput=1

\documentclass[11pt]{article}

\usepackage[]{acl}
\usepackage{times}
\usepackage{latexsym}
\usepackage{booktabs}

\usepackage[T1]{fontenc}

\usepackage[utf8]{inputenc}

\usepackage{microtype}
\input{math_commands.tex}

\makeatletter

\newcommand*\totht[1]{\dimexpr\ht#1+\dp#1\relax}
\newcommand*\leading{{\setbox0\hbox{\strut}\the\totht0}}
\newcommand*\fntsize{{\setbox0\hbox{Mg}\the\totht0}}
\newcommand*\showsize[1]{{#1 {\ttfamily\string#1} (\f@size pt) \fntsize/\leading}\par}
\makeatother

\title{DAMO-NLP at SemEval-2022 Task 11:\\ A Knowledge-based System for Multilingual Named Entity Recognition}

\usepackage{microtype}
\usepackage{bbm}
\usepackage{amsmath}
\usepackage{amsfonts}
\usepackage{algorithm}
\usepackage{algpseudocode}
\usepackage{booktabs}
\usepackage{xparse}
\usepackage{graphicx}
\usepackage{multicol}
\usepackage{multirow}
\usepackage{tikz}
\usetikzlibrary{shapes,arrows}
\usepackage{natbib}
\usepackage{caption}
\usepackage{subcaption}
\usepackage{tikz-dependency}
\usepackage{wrapfig}
\usepackage{pgfplots}
\pgfplotsset{compat=1.17} 
\usepackage[T1]{fontenc}
\usepackage{subfiles}
\usepackage{readarray}
\usepackage{xcolor}
\usepackage{import}
\usepackage{hhline}
\usepackage{enumitem}
\usepackage{boldline}
\usepackage{mathtools}
\usepackage{colortbl}

\newcommand{\mcL}{\mathcal{L}}

\newcommand{\Wvec}{\mathbf{W}}

\newcommand{\bvec}{\mathbf{b}}

\author{Xinyu Wang$^{\diamond\star}$, Yongliang Shen$^{\spadesuit\star}$, Jiong Cai$^{\diamond\star}$, Tao Wang, Xiaobin Wang$^\dagger$, Pengjun Xie$^\dagger$\\
\textbf{Fei Huang$^\dagger$, Weiming Lu$^\spadesuit$, Yueting Zhuang$^\spadesuit$, Kewei Tu$^\diamond$, Wei Lu$^\ddagger$, Yong Jiang$^\dagger$\thanks{\;\,: project lead. $^\star$: equal contributions.}} \\
$^\dagger$DAMO Academy, Alibaba Group \\
$^\diamond$School of Information Science and Technology, ShanghaiTech University \\
$^\spadesuit$College of Computer Science and Technology, Zhejiang University \\
$^\ddagger$StatNLP Research Group, Singapore University of Technology and Design \\
{\tt \{wangxy1,caijiong,tukw\}@shanghaitech.edu.cn}\\ 
{\tt\{syl,luwm\}@zju.edu.cn, luwei@sutd.edu.sg}\\
{\tt yongjiang.jy@alibaba-inc.com}}

\begin{document}
\maketitle
\begin{abstract}
The MultiCoNER shared task aims at detecting semantically ambiguous and complex named entities in short and low-context settings for multiple languages. The lack of contexts makes the recognition of ambiguous named entities challenging. To alleviate this issue, our team \textbf{DAMO-NLP} proposes a knowledge-based system, where we build a multilingual knowledge base based on Wikipedia to provide related context information to the named entity recognition (NER) model. Given an input sentence, our system effectively retrieves related contexts from the knowledge base. The original input sentences are then augmented with such context information, allowing significantly better contextualized token representations to be captured. Our system wins 10 out of 13 tracks in the MultiCoNER shared task.\footnote{Our code is publicly available at \url{https://github.com/Alibaba-NLP/KB-NER}.}


\end{abstract}

\section{Introduction}
\label{intro}

\begin{figure}[t]
	\centering
	\includegraphics[scale=0.26]{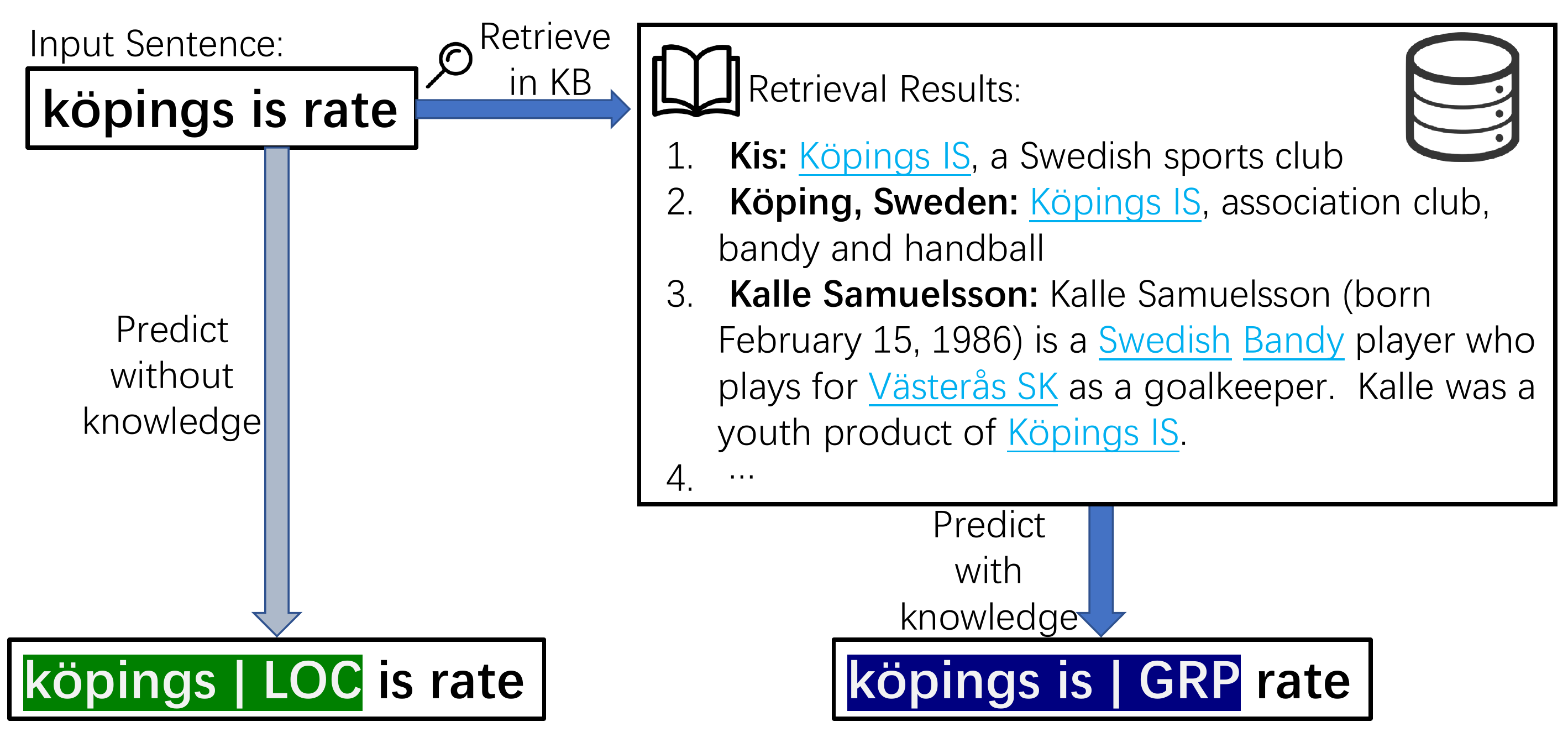}
	\caption{A motivating example from the English test set. In the retrieval results, the bold phrases are the title of the retrieved page and the underlined phrases contain the hyperlinks to other pages. LOC and GRP are entity labels representing location and group respectively.}
	\label{fig:motivate}
\end{figure}

The MultiCoNER shared task \citep{multiconer-report} aims at building Named Entity Recognition (NER) systems for 11 languages, including English, Spanish, Dutch, Russian, Turkish, Korean, Farsi, German, Chinese, Hindi, and Bangla. The task has three kinds of tracks including one multilingual track, 11 monolingual tracks and one code-mixed track. The multilingual track requires training multilingual NER models that are able to handle all languages. The monolingual tracks require training individual monolingual models where each model works for only one language. The code-mixed track requires handling code-mixed samples (sentences that may involve multiple languages). The datasets mainly contain sentences from three domains: Wikipedia, web questions and user queries, which are usually short and low-context sentences. Moreover, these short sentences usually contain semantically ambiguous and complex entities, which makes the problem more difficult.
In practice, professional annotators usually use their domain knowledge to disambiguate such kinds of entities. They may retrieve the related documents from a knowledge base (KB) or from a search engine to better guide them the annotation of ambiguous named entities \citep{wang-etal-2019-crossweigh}.
Therefore, we believe retrieving related knowledge can help the NER model to disambiguate hard samples in the shared task as well. A motivating example is shown in Figure \ref{fig:motivate}, which shows how the retrieval results could help to improve the prediction in practice.


In this paper, we propose a general knowledge-based system for the MultiCoNER shared task. We propose to retrieve the related documents of the input sentence so that the recognition of difficult entities can be significantly eased. Based on Wikipedia of the 11 languages, we build a multilingual KB to search for the related documents of the input sentence. We then feed the input sentence and the related documents into the NER model. Moreover, we propose an iterative retrieval approach to improve the retrieval quality.  During training, we propose multi-stage fine-tuning. We first train a multilingual model so that the NER model can learn from all annotations. Next, we train the monolingual models (one for each language) and a code-mixed model by using the fine-tuned XLM-RoBERTa (XLM-R) \citep{conneau-etal-2020-unsupervised} embeddings in the multilingual model as initialization to further boost model performance on monolingual and code-mixed tracks. For each track, we train multiple models with different random seeds and use majority voting to form the final predictions. 

Besides the system description, we make the following observations based on our experiments:
\setlist{nolistsep}
\begin{enumerate}[leftmargin=*,noitemsep]
    \item Knowledge-based systems can significantly improve both in- and out-of-domain performance compared with system without knowledge inputs.
    \item Our multi-stage fine-tuning approach can help improve model performance in all the monolingual and code-mixed tracks. The approach can also reduce the training time to speed up our system building at different stages.
    \item Our iterative retrieval strategy can further improve the retrieval quality and result in significant improvement on the performance of code-mixed track. 
    \item Searching over Wikipedia KB performs better than using online search engines on the MultiCoNER datasets.
    \item Comparing with other model variants we have tried, our NER model enjoys a good balance between model performance and speed.
\end{enumerate}





\section{Related Work}
NER \citep{Sundheim1995NamedET} is a fundamental task in natural language processing. The task has a lot of applications in various domains such as social media \citep{derczynski-etal-2017-results}, news \cite{tjong-kim-sang-2002-introduction,tjong-kim-sang-de-meulder-2003-introduction}, E-commerce \citep{10.1145/3404835.3463102,wang-etal-2021-improving}, and medical domains \citep{dougan2014ncbi,li2016biocreative}. Recently, pretrained contextual embeddings such as BERT \citep{devlin-etal-2019-bert}, XLM-R and LUKE \citep{yamada-etal-2020-luke} have significantly improved the NER performance. The embeddings are trained on large-scale unlabeled data such as Wikipedia, which can significantly improve the contextual representations of named entities. Recent efforts \citep{peters-etal-2018-deep,akbik-etal-2018-contextual,strakova-etal-2019-neural} concatenate different kinds of pretrained embeddings to form stronger token representations. Moreover, the embeddings are trained over long documents, which allows the model to easily model long-range dependencies to disambiguate complex named entities in the sentence. Recently, a lot of work shows that utilizing the document-level contexts in the CoNLL NER datasets can significantly improve token representations and achieves state-of-the-art performance \citep{yu-etal-2020-named,luoma-pyysalo-2020-exploring,yamada-etal-2020-luke,wang2020automated}. 
However, the lack of context in the MultiCoNER datasets means the embeddings cannot take advantage of long-range dependencies for entity disambiguation.
Recently, \citet{wang-etal-2021-improving} use Google search to retrieve external contexts of the input sentence and successfully achieve state-of-the-art performance across multiple domains. We adopt this idea so that the embeddings can utilize the related knowledge by taking the advantage of long-range dependencies to form stronger token representations. 
Comparing with \citet{wang-etal-2021-improving}, we build the local KB based on Wikipedia because the KB matches the in-domain data of the shared task and is fast enough to meet the time requirement in the test phase\footnote{There are only 7 days for the test phase.}.

Fine-tuning pretrained contextual embeddings is a useful and effective approach to many NLP tasks. Recently, some of the research efforts propose to further train the fine-tuned embeddings with specific training data or in a larger model architecture to improve model performance.  \citet{shi-lee-2021-tgif} proposed two-stage fine-tuning, which first trains a general multilingual Enhanced Universal Dependency \citep{bouma-etal-2021-raw} parser and then fine-tunes on each specific language separately. \citet{wang2020automated} proposed to train models through concatenating fine-tuned embeddings. We extend these ideas as multi-stage fine-tuning, which improves the accuracy of monolingual models that use fine-tuned multilingual embeddings as initialization in training. Moreover, multi-stage fine-tuning can accelerate the training process in system building.







\section{Our System}

\begin{figure}[t!]
	\centering
	\includegraphics[scale=0.33]{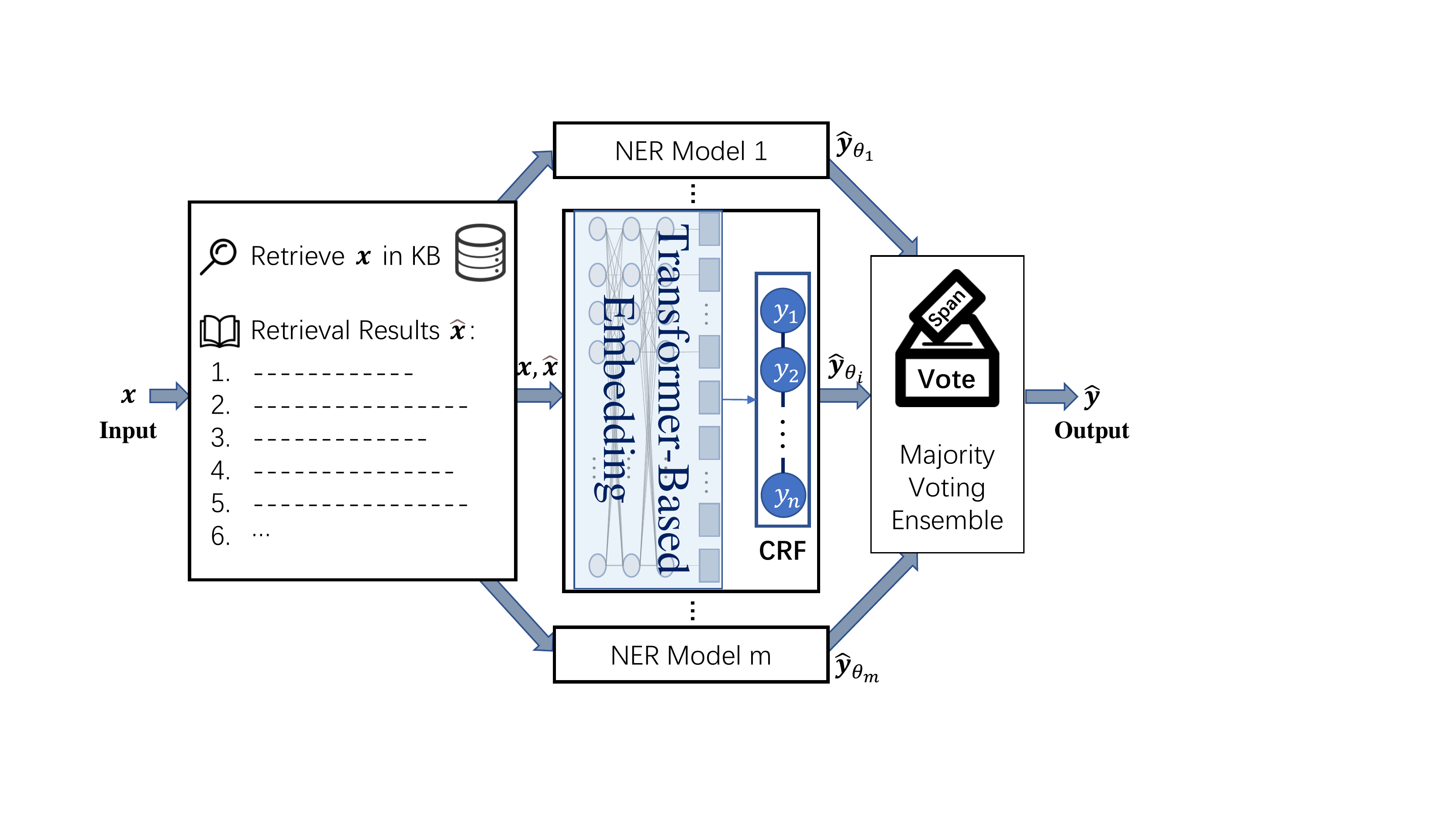}
	\caption{The architecture of our knowledge-based NER system.}
	\label{fig:architecture}
\end{figure}

We introduce how our knowledge-based NER system works in this section. Given a sentence of $n$ tokens $\vx = \{x_1, \cdots, x_n\}$, the sentence is fed into our knowledge retrieval module. The knowledge retrieval module takes the sentence as the query and retrieves top-$k$ related paragraphs in KB. The system then concatenates the input sentence and the related paragraphs together and feeds the concatenated sequence into the embeddings. The output token representations of the input sentence are fed into a linear-chain conditional random field (CRF) \citep{10.5555/645530.655813} layer and the CRF layer produces the label predictions. Given the label predictions of multiple NER models with different random seeds, the ensemble module uses a voting strategy to decide the final predictions $\hat{\vy} = \{\hat{y}_1, \cdots, \hat{y}_n\}$ of the sentence. The architecture of our framework is shown in Figure \ref{fig:architecture}.

\subsection{Knowledge Retrieval Module}

Retrieval-augmented context is effective for named entity recognition tasks \citep{wang-etal-2021-improving}, as external relevant contexts can provide auxiliary information for disambiguating complex named entities. We construct multilingual KBs based on Wikipedia pages of the 11 languages, and then retrieve relevant documents by using the input sentence as a query. These retrieved documents act as contexts and are fed into the NER module. To enhance the retrieval quality, we further designed an iterative retrieval approach, which incorporates predicted entities of NER models into the search query.

\paragraph{Knowledge Base Building}
\label{build}

Wikipedia is an evolving source of knowledge that can facilitate many NLP tasks \citep{chen-etal-2017-reading, verlinden-etal-2021-injecting}. Wikipedia provides a rich collection of mention hyperlinks (referred to as wiki anchors). For example, in the sentence ``\textit{Steve Jobs founded Apple}'', entities ``\textit{Steve Jobs}'' and ``\textit{Apple}'' are linked to the wiki entries \texttt{Steve\_Jobs} and \texttt{Apple\_Inc} respectively. For the NER task, these anchors provide useful clues on where the entities are to the model. Based on Wikipedia we can build local Wikipedia search engines to retrieve the relevant context of the input sentences for each language. 

We download the latest (2021-12-20) version of the Wikipedia dump from Wikimedia\footnote{\url{https://dumps.wikimedia.org/}} and convert it to plain texts. Then we use ElasticSearch (ES)\footnote{\url{https://www.elastic.co/}} to index them. ElasticSearch is document-oriented, and the document is the least searchable unit. We define the document in our local Wikipedia search engines with three fields: sentence, paragraph and title. We create inverted indexes on both the sentence field and the title field. 
The former is used as a sentence-level full-text retrieval field, while the latter indicates the core entity described by the wiki page and can be used as an entity-level retrieval field. 
The paragraph field stores the contexts of the sentence. To take advantage of the rich wiki anchors in Wikipedia paragraphs, we marked them with special markers. 
For example, to incorporate the hyperlinks [\textit{Apple} $\rightarrow$ \texttt{Apple Inc}] and [\textit{Steve Jobs} $\rightarrow$ \texttt{Steve Jobs}] to the paragraph, we transformed ``\textit{Steve Jobs founded Apple}'' into ``\textit{<e:Steve Jobs>Steve Jobs</e> founded <e:Apple\_inc>Apple</e>}''\footnote{<e:XXX>YYY</e>: where XXX is the title of the linked page and YYY is the phrase with hyperlink in the sentence.}. 



\paragraph{Sentence Retrieval}
\label{sec:sentence}
Retrieval at the sentence level takes the input sentence as a query and retrieves the top-$k$ documents on the sentence field.
Given an input sentence, we select the corresponding search engine according to the language of the sentence. 



\paragraph{Iterative Entity Retrieval}
\label{sec:coarse2fine}
The core of the NER task lies in the entities, while retrieval at the sentence level overlooks the key entities in the sentences. For this reason, we consider the relevance of the entities in the sentence to the title field in the documents during retrieval. We concatenate the entities in the sentences with ``|'' and then retrieve them on the title field. On the training and development sets, we utilize the ground-truth entities directly. On the test set, we first perform the sentence retrieval and then use the entity mentions\footnote{Here we define mentions as the named entities ignoring the entity types.} predicted by the model for entity retrieval.
This bootstrapping manner can be applied for $T$ turns. 

\paragraph{Context Processing}
\label{sec:post-processing}
After top-$k$ results from the KB are retrieved, the system post-processes the retrieved documents into the contexts of the input sentence. There are three options of utilizing the texts in the documents, which are: 1) use the matched paragraph; 2) use the matched sentence; 3) use the matched sentence but remove the wiki anchors. We compare the performance of each option in section \ref{sec:context_type}. In each retrieved document, we concatenate the title and texts together to form the context $\hat{\vx}_i$. The results are then concatenated into $\{\hat{\vx}_1, \cdots, \hat{\vx}_k\}$ based on the retrieval ranking.

\subsection{Named Entity Recognition Module}
In our system, we use XLM-R large as the embedding for all the tracks. It is a multilingual model and is applicable to all tracks. Given the input sentence $\vx$ and the retrieved contexts $\{\hat{\vx}_1, \cdots, \hat{\vx}_k\}$, we add the separator token (i.e., ``</s>'' in XLM-R) between them and concatenated them together to form the input $\tilde{\vx}$ of the NER module. We chunk retrieved texts to avoid the amount of subtoken in the sequence exceeding the maximum subtoken length in XLM-R (i.e.,  512 in XLM-R).

Our system regards the NER task as a sequence labeling problem. The embedding layer in the NER module encode the concatenated sequence $\tilde{\vx}$ and output the corresponding token representations $\{\vv_1, \cdots, \vv_n,\cdots\}$. The module then feeds the token representations $\{\vv_1, \cdots, \vv_n\}$ of the input sentence into a linear-chain CRF layer to obtain the conditional probability $p_\theta(\vy|\tilde{\vx})$:
\begin{align}
    \psi(y', y, \vv_i) &= \exp(\Wvec_{y}^{T} \vv_i + \bvec_{y',y}) \label{eq:psi}\\
    p_\theta(\vy|\tilde{\vx}) &= \frac{\prod\limits_{i=1}^{n} \psi(y_{i-1}, y_i, \vv_i)}{\sum\limits_{\vy' \in \mathcal{Y}(\vx)} \prod\limits_{i=1}^{n} \psi(y'_{i-1}, y'_i, \vv_i)}\nonumber
\end{align}
where $\theta$ represents the model parameters and $\mathcal{Y}(\vx)$ denotes the set of all possible label sequences given $\vx$. In the potential function $\psi(y', y, \vv_i)$, $\Wvec_{y}^{T} \vv_i$ is the emission score and $\bvec_{y',y}$ is the transition score, where $\Wvec^{T}\in \sR^{t\times d}$ and $\bvec \in \sR^{t \times t}$ are parameters and the subscripts ${y'}$ and ${y}$ are the indices of the matrices. During training, the negative log-likelihood loss $\mcL_{\text{NLL}}(\theta) = - \log p_\theta(\vy^*|\tilde{\vx})$ for the concatenated input sequence with gold labels $\vy^*$ is used.
During inference, the model prediction $\hat{\vy}_{\theta}$ is given by Viterbi decoding.

\subsection{Ensemble Module}
\label{sec:ensemble}

Given predictions $\{\hat{\vy}_{\theta_1}, \cdots, \hat{\vy}_{\theta_m}\}$ from $m$ models with different random seeds, we use majority voting to generate the final prediction $\hat{\vy}$. We convert the label sequences into entity spans to perform majority voting. Following \citet{yamada-etal-2020-luke}, the module ranks all spans in the predictions by the number of votes in descending order and selects the spans with more than 50\% votes into the final prediction. The spans with more votes are kept if the selected spans have overlaps and the longer spans are kept if the spans have the same votes.

\section{Experimental Setup}
\subsection{Data and Evaluation Methodology}
We use the official MultiCoNER dataset \citep{multiconer-data} in all tracks to train our NER models. There are mainly three domains in the dataset: LOWNER (Low-Context Wikipedia NER) contains low-context sentences from Wikipedia; MSQ (MS-MARCO Question NER) is based on MS-MARCO web question corpus \citep{nguyen2016ms} containing a lot of natural language questions; ORCAS (Search Query NER) contains user queries from Microsoft Bing \citep{craswell2020orcas}. The MSQ and ORCAS samples are taken as out-of-domain data in the shared task. The training and development sets only contain a small collection of samples of these two domains and mainly contain data from the LOWNER domain. The test set, however, contains much more MSQ and ORCAS samples to assess the out-of-domain performance.

The results of the shared task are evaluated with the entity-level macro F1 scores, which treat all the labels equally. In comparison, most of the publicly available NER datasets (e.g., CoNLL 2002, 2003 datasets) are evaluated with the entity-level micro F1 scores, which emphasize common labels.



\begin{table*}[ht!]
\centering
\scalebox{0.75}{
\begin{tabular}{lccccccccccccc|c}
\toprule
System &		\textsc{\textbf{en}}&	\textsc{\textbf{es}}&	\textsc{\textbf{nl}}&	\textsc{\textbf{ru}}&	\textsc{\textbf{tr}}&	\textsc{\textbf{ko}}&	\textsc{\textbf{fa}}&	\textsc{\textbf{de}}&	\textsc{\textbf{zh}}&	\textsc{\textbf{hi}}&	\textsc{\textbf{bn}}&	\textsc{\textbf{multi}}&	\textsc{\textbf{mix}}&	\textsc{\textbf{Avg.}} \\
\midrule
Ours: Baseline&	77.81 &	76.80 &	80.51 &	74.65 &	72.83 &	70.81 &	72.68 &	81.92 &	65.56 &	67.80 &	65.27 &	74.19 &	77.75 &	73.74 \\
Sliced & 74.54 &75.11 &77.66 &73.73 &68.77 &70.66 &68.66 &78.90 &65.21 &67.00 &63.05 &71.07 &72.74 &71.32 \\
RACAI & 75.78 &75.62 &78.41 &74.60 &70.42 &71.74 &70.42 &79.39 &62.70 &68.08 &66.28 &72.10 &79.37 &72.69 \\
USTC-NELSLIP &	85.47 &	85.44 &	87.67 &	83.82 &	85.52 &	86.36 &	87.05 &	89.05 &	\textbf{81.69} &	84.64 &	\textbf{84.24} &	85.30 &	\textbf{92.90} &	86.09 \\
Ours&	\textbf{91.22} &	\textbf{89.94} &	\textbf{90.50} &	\textbf{91.50} &	\textbf{88.69} &	\textbf{88.59} &	\textbf{89.70} &	\textbf{90.65} &	78.06 &	\textbf{86.23} &	83.51 &	\textbf{85.31} &	91.79 &	\textbf{88.13} \\
\bottomrule
\end{tabular}
}
\caption{Part of the official results and the post-evaluation results of our baseline system. }
\label{tab:results}
\end{table*}

\subsection{Training}

\paragraph{NER Model Training}
Before building the final system, we compare a lot of variants of the system. We train these variant models on the training set for 3 times each with different random seeds and compare the averaged performance of the models. According to the dataset sizes, we train the models for 5 epochs, 10 epochs and 100 epochs for multilingual, monolingual and code-mixed models respectively. Our final NER models are trained on the combined dataset including both the training and development sets on each track to fully utilize the labeled data. For models trained on the training set, we use the best macro F1 on the development set during training to select the best model checkpoint. For models trained on the combined dataset, we use the final model checkpoint after training\footnote{Please refer to Appendix \ref{app:exp} for detailed settings.}.

\paragraph{Multi-stage Fine-tuning}
Besides our final settings, we have a lot of stages of KB settings during our system building.
Multi-stage fine-tuning aims at transferring the parameters of fine-tuned embeddings in a model at an early stage into other models in the next stage. The approach stores the checkpoint of fine-tuned XLM-R embeddings at the early stage and uses it as the initialization of XLM-R embeddings for model training at the next stage. One benefit of multi-stage fine-tuning is the monolingual and code-mixed models,  can utilized the annotations of all the tracks to further improve the model performance. XLM-R embedding is a multilingual embedding with strong cross-lingual transferability over all 11 languages. Therefore, we use the checkpoint of fine-tuned multilingual model for continue fine-tuning on the monolingual and code-mixed models. Another benefit of multi-stage fine-tuning is that it accelerates the training speed. As the size of the multilingual dataset is relatively large, it is quite time-consuming to train a multilingual model. When we try different types of KB, we can utilize the checkpoints of multilingual models at the previous stage to train the monolingual and code-mixed models with new types of contexts without training new multilingual models. Moreover, we can reduce the training epochs for faster speed since the XLM-R checkpoints have already learned from all the datasets.







\paragraph{Continue Pretraining}
To make XLM-R learn the data distribution of the shared task, we combine the training and development sets on the monolingual tracks to build a corpus to continue pretrain XLM-R. Specifically, we collocate all sentences according to their languages, then cut the text into chunks of fixed length, and train the model on these text chunks using the Masked Language Modeling objective. We continue pretrain XLM-R for 5 epochs. We use the continue pretrained XLM-R model as the initialization of the multilingual models during training.

\section{Results and Analysis}
In this section, we use language codes\footnote{\url{https://en.wikipedia.org/wiki/List_of_ISO_639-1_codes}} to represent languages, and use \textsc{\textbf{multi}} and \textsc{\textbf{mix}} to represent multilingual and code-mixed tracks respectively\footnote{Please refer to Appendix \ref{app:more_anal} for more analysis.}.

\subsection{Main Results}
\label{sec:results}
There are 55 teams that participated in the shared task. Due to limited space, we only compare our system with the systems from teams USTC-NELSLIP, RACAI and Sliced\footnote{Please refer to \url{https://multiconer.github.io/results} for more details about the results.}. In the post-evaluation phase, we evaluate a baseline system without using the knowledge retrieval module to further show the effectiveness of our knowledge-based system. The official results and the results of our baseline system are shown in Table \ref{tab:results}. Our system performs the best on 10 out of 13 tracks and is competitive on the other 3 tracks. Moreover, our system outperforms our baseline by 14.39 F1 on average, which shows the knowledge retrieval module is extremely helpful for disambiguating complex entities leading to significant improvement on model performance.

\begin{table*}[ht!]
\small
\setlength\tabcolsep{5.5pt}
\centering
\begin{tabular}{cclccccccccccc|c}
\toprule
&&&	\textsc{\textbf{en}}&	\textsc{\textbf{es}}&	\textsc{\textbf{nl}}&	\textsc{\textbf{ru}}&	\textsc{\textbf{tr}}&	\textsc{\textbf{ko}}&	\textsc{\textbf{fa}}&	\textsc{\textbf{de}}&	\textsc{\textbf{zh}}&	\textsc{\textbf{hi}}&	\textsc{\textbf{bn}}&	\textsc{\textbf{Avg.}} \\
\midrule
\parbox[t]{2mm}{\multirow{3}{*}{\rotatebox[origin=c]{90}{\scriptsize{In-domain}}}}
&
\parbox[t]{2mm}{\multirow{3}{*}{\rotatebox[origin=c]{90}{\scriptsize{\textsc{lowner}}}}}
&
Baseline&	88.70 &	86.54 &	89.92 &	81.52 &	88.52 &	86.25 &	81.85 &	91.71 &	85.43 &	83.13 &	82.69 &	86.02 \\
&&Ours&	96.78 &	96.19 &	97.96 &	96.60 &	96.43 &	96.83 &	96.48 &	94.89 &	88.66 &	84.18 &	86.31 &	93.76 \\
&&$\Delta$& +8.08 & +9.65 & +8.04 & \llap{+}15.08 & +7.91 & \llap{+}10.58 & \llap{+}14.63 & +3.18 & +3.23 & +1.05 & +3.62 & +7.74 \\
\midrule
\parbox[t]{2mm}{\multirow{6}{*}{\rotatebox[origin=c]{90}{\scriptsize{{Out-of-domain}}}}}
&\parbox[t]{2mm}{\multirow{3}{*}{\rotatebox[origin=c]{90}{\scriptsize\textsc{msq}}}}
&Baseline&	70.49 &	71.86 &	72.63 &	72.31 &	75.49 &	68.57 &	71.54 &	74.63 &	67.38 &	73.57 &	58.66 &	70.65 \\
&&Ours&	83.50 &	83.10 &	83.34 &	87.03 &	88.76 &	81.96 &	87.36 &	86.18 &	79.80 &	89.20 &	72.00 &	83.84 \\
&&$\Delta$& \llap{+}13.01 &	\llap{+}11.24 & \llap{+}10.71 & \llap{+}14.72 & \llap{+}13.27 & \llap{+}13.39 & \llap{+}15.82 & \llap{+}11.55 & \llap{+}12.42 & \llap{+}15.63 & \llap{+}13.34 & \llap{+}13.19 \\
\cmidrule{2-15}
&\parbox[t]{2mm}{\multirow{3}{*}{\rotatebox[origin=c]{90}{\scriptsize\textsc{orcas}}}}
&Baseline&	62.07 &	62.71 &	67.39 &	64.83 &	66.92 &	56.08 &	65.52 &	67.52 &	55.34 &	62.03 &	60.68 &	62.83 \\
&&Ours&	83.72 &	81.33 &	80.29 &	85.00 &	85.85 &	81.06 &	84.84 &	84.40 &	72.11 &	85.75 &	82.13 &	82.41 \\
&&$\Delta$& \llap{+}21.65 & \llap{+}18.62 & \llap{+}12.90 & \llap{+}20.17 & \llap{+}18.93 & \llap{+}24.98 & \llap{+}19.32 & \llap{+}16.88 & \llap{+}16.77 & \llap{+}23.72 & \llap{+}21.45 & \llap{+}19.58 \\
\bottomrule
\end{tabular}
\caption{Per-domain macro F1 score on the test set of our system and our baseline system for each language. $\Delta$ represents the relative improvements of our system over the baseline system. }
\label{tab:domain_f1}
\end{table*}

\subsection{How Significant is the Role of Knowledge-based System on Each Domain?}
To further show the effectiveness of our knowledge-based system, we show the relative improvements of our system over our baseline system on each domain in Table \ref{tab:domain_f1}. We observe that in most of the cases, the two out-of-domain test sets have more relative improvements than the in-domain test set. This observation shows that the knowledge from Wikipedia can not only improve the performance of the LOWNER domain which is the same domain as the KB, but also has very strong cross-domain transferability to other domains such as web questions and user queries. According to the baseline performance over the three domains, the ORCAS domain has the lowest score, which shows the challenges in recognizing named entities in user queries. However, our retrieved documents in KB can significantly ease the challenges in this domain and results in the highest improvement out of the three domains.

\subsection{How Relevant Are the Retrieval Results to the Queries?}

\begin{figure*}[t!]
  \centering
  \includegraphics[width=\linewidth]{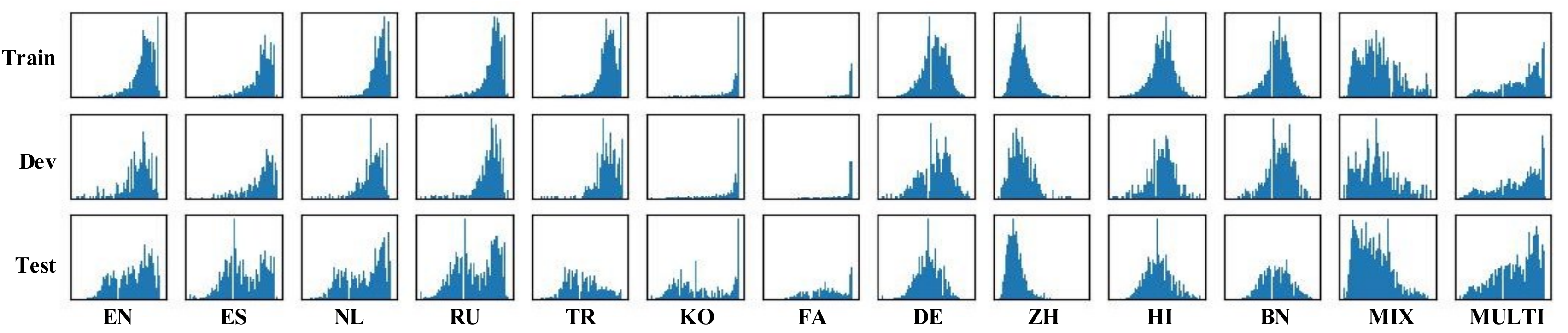}
  \caption{The distribution of the character-level IoU between the query and its top-1 result. Each subplot is a histogram on the corresponding dataset, where the \textit{x}-axis indicates the IoU values ranging from 0 to 1.}
  \label{fig:iou}
\end{figure*}

To evaluate the relevance of the retrieval results to the query, we define a character-level relevance metric, which calculates the Intersection-over-Union (IoU) between the characters of query and result. 
Assuming that the character sets\footnote{The sets take repeat characters as different characters.} of query and retrieval result are $A$ and $B$ respectively, then the character-level IoU is $\frac{A\cap B}{A\cup B}$.
We calculate the character-level IoU of the sentence and its top-1 retrieval result on all tracks, and plot its distribution on the training, development and test set in Figure \ref{fig:iou}. We have the following observations: 
1) the IoU values are concentrated around 1.0 on the training and development sets of \textsc{\textbf{en}}, \textsc{\textbf{es}}, \textsc{\textbf{nl}}, \textsc{\textbf{ru}}, \textsc{\textbf{tr}}, \textsc{\textbf{ko}}, \textsc{\textbf{fa}}, which indicates that most of the samples were derived from Wikipedia.
Therefore, by retrieving, we can obtain the original documents for these samples. 
2) the distribution of data on the test set is consistent with the training and development sets for most languages, except for \textsc{\textbf{tr}}.
On \textsc{\textbf{tr}}, the character-level IoU values of the samples and query results cluster at around 0.5. We hypothesize that this is because the source of the test set for \textsc{\textbf{tr}} is different from the training set.
However, the model still performs strongly on this language, suggesting that the model can mitigate the difficulties caused by inconsistent data distribution by retrieving the context from Wikipedia.

\begin{table*}[ht!]
\small
\setlength\tabcolsep{4pt}
\centering
\begin{tabular}{lccccccccccccc}
\toprule
&	\textsc{\textbf{en}}&	\textsc{\textbf{es}}&	\textsc{\textbf{nl}}&	\textsc{\textbf{ru}}&	\textsc{\textbf{tr}}&	\textsc{\textbf{ko}}&	\textsc{\textbf{fa}}&	\textsc{\textbf{de}}&	\textsc{\textbf{zh}}&	\textsc{\textbf{hi}}&	\textsc{\textbf{bn}}&	\textsc{\textbf{multi}} &	\textsc{\textbf{mix}}\\
\midrule
Baseline&	87.13 &	85.88 &	88.87 &	82.38 &	86.22 &	85.98 &	81.25 &	91.21 &	87.65 &	82.62 &	82.80 &	85.78 &	77.92  \\
Google Search&	92.46 &	88.68 &	91.58 &	85.88 &	89.83 &	88.95 &	82.96 &	93.56 &	89.16 &	84.27 &	84.38  &	87.84 &	86.26 \\
Wiki-\textsc{Para}&	\textbf{95.82} & 94.19 &	\textbf{97.53} &	95.53 &	\textbf{97.40} &	96.05 &	\textbf{95.93} &	92.83 &	87.10 &	82.78 &	83.35 &	93.51 &	85.16  \\
Wiki-\textsc{Sent}&	87.62 &	89.33 &	92.90 &	79.41 &	89.00 &	91.49 &	95.99 &	94.42 &	\textbf{89.47} & 84.55 &	84.12 &	89.34 &	78.65  \\
Wiki-\textsc{Sent$_{\text{-link}}$}&	86.83 &	87.65 &	91.86 &	79.15 &	86.66 &	86.36 &	84.37 &	94.46 &	89.32 &	84.78 &	84.83 &	87.35 &	80.07  \\
Wiki-\textsc{Para}+\textsc{Iter}$_G$&	94.89 &	\textbf{94.44} &	97.45 &	\textbf{95.59} &	96.89 &	\textbf{96.34} &	95.83 &	\textbf{94.62} &	88.47 &	\textbf{86.43} &	\textbf{85.85} &	\textbf{93.60}  &	\textbf{90.52} \\
\bottomrule
\end{tabular}
\caption{A comparison of the models utilizing different types of knowledge on the \textbf{development set}. }
\label{tab:context_comparison}
\end{table*}

\begin{table}[ht!]
\small
\centering
\begin{tabular}{lccc}
\toprule
&	\textsc{\textbf{hi}}&	\textsc{\textbf{bn}}&	\textsc{\textbf{mix}}\\
\midrule
 Wiki-\textsc{Para}+\textsc{Iter}$_G$ &	86.43 &	85.85 &	90.52  \\
 Wiki-\textsc{Sent}+\textsc{Iter}$_G$ &	85.69 &	86.57 &	91.38 \\
 Wiki-\textsc{Sent$_{\text{-link}}$}+\textsc{Iter}$_G$ &	86.15 &	86.13 &	91.38 \\
\midrule
 Wiki-Opt$_\text{Best}$ &	\textbf{84.78} &	\textbf{84.83} &	85.16 \\
 Wiki-Opt$_\text{Best}$+\textsc{Iter}$_P$ &	83.36 &	84.37 &	\textbf{88.97} \\
\bottomrule
\end{tabular}
\caption{Effectiveness of iterative retrieval. Opt$_\text{Best}$ represents using the best context option for each language.}
\label{tab:context_coarse2fine}
\end{table}

\begin{table}[ht!]
\small
\centering
\begin{tabular}{lccc}
\toprule
&	\textsc{\textbf{hi}}&	\textsc{\textbf{bn}}&	\textsc{\textbf{mix}}\\
\midrule
Wiki-Opt$_\text{Best}$&	90.02&	90.81&	96.72 \\ 
Wiki-Opt$_\text{Best}$-Mention&	90.76&	90.75&	96.71 \\ 
\bottomrule
\end{tabular}
\caption{A comparison of mention detection F1 score over NER models and mention detection models.}
\label{tab:mention}
\end{table}

\begin{table}[t!]
\small
\setlength\tabcolsep{4pt}
\centering
\begin{tabular}{lr}
\toprule
Module &	Sentences/Second\\
\midrule
Local Knowledge Base Retrieval & 64.52\\
Google Search Retrieval & 1.50\\
NER Module - Training & 2.91 \\
NER Module - Prediction & 8.13 \\
\bottomrule
\end{tabular}
\caption{Model speed of the knowledge retrieval module and NER module in our system. }
\label{tab:speed}
\end{table}


\begin{table*}[ht!]
\small
\setlength\tabcolsep{5pt}
\centering
\begin{tabular}{lcccccccccccc|c}
\toprule
&\textsc{\textbf{en}}&	\textsc{\textbf{es}}&	\textsc{\textbf{nl}}&	\textsc{\textbf{ru}}&	\textsc{\textbf{tr}}&	\textsc{\textbf{ko}}&	\textsc{\textbf{fa}}&	\textsc{\textbf{de}}&	\textsc{\textbf{zh}}&	\textsc{\textbf{hi}}&	\textsc{\textbf{bn}}&	\textsc{\textbf{mix}}&	\textsc{\textbf{Avg.}} \\
\midrule
XLM-R&	92.46 &	88.68 &	91.58 &	85.88 &	89.83 &	88.95 &	82.96 &	93.56 &	89.16 &	84.27 &	84.38 &	84.52 &	88.02  \\
CE&	\textbf{92.49} &	\textbf{88.97} &	\textbf{92.20} &	\textbf{86.21} &	\textbf{90.47} &	\textbf{89.01} &	\textbf{83.53} &	\textbf{93.96} &	\textbf{89.40} &	\textbf{84.86} &	\textbf{85.38} &	\textbf{87.35} &	\textbf{88.65}  \\
\bottomrule
\end{tabular}
\caption{A comparison of CE models and XLM-R models. Both kinds of models utilize the knowledge from Google Search. The scores are the averaged macro F1 score on the \textbf{development set}.}
\label{tab:google_ce}
\end{table*}

\begin{table*}[t!]
\small
\centering
\scalebox{0.92}{
\begin{tabular}{lcccccccccccc|c}
\toprule
&	\textsc{\textbf{en}}&	\textsc{\textbf{es}}&	\textsc{\textbf{nl}}&	\textsc{\textbf{ru}}&	\textsc{\textbf{tr}}&	\textsc{\textbf{ko}}&	\textsc{\textbf{fa}}&	\textsc{\textbf{de}}&	\textsc{\textbf{zh}}&	\textsc{\textbf{hi}}&	\textsc{\textbf{bn}}&	\textsc{\textbf{mix}}&	\textsc{\textbf{Avg.}}\\
\midrule
Baseline w/ MF&	\textbf{87.13} &	\textbf{85.88} &	\textbf{88.87} &	\textbf{82.38} &	\textbf{86.22} &	\textbf{85.98} &	\textbf{81.25} &	\textbf{91.21} &	\textbf{87.65} &	\textbf{82.62} &	\textbf{82.80} &	\textbf{77.92} &	\textbf{84.99}  \\
Baseline w/o MF&	85.88 &	84.28 &	87.98 &	81.01 &	84.61 &	83.98 &	79.98 &	89.54 &	85.57 &	79.90 &	81.18 &	68.21 &	82.68  \\
\bottomrule
\end{tabular}
}
\caption{A comparison of training the NER models with and without multi-stage fine-tuning (MF) for our baseline system on the \textbf{development set}. }
\label{tab:continue_tuning}
\end{table*}

\begin{table}[ht!]
\small
\setlength\tabcolsep{2pt}
\centering
\scalebox{0.92}{
\begin{tabular}{lccccccc|c}
\toprule
&		\textsc{\textbf{en}}&	\textsc{\textbf{es}}&	\textsc{\textbf{nl}}&	\textsc{\textbf{ru}}&	\textsc{\textbf{tr}}&	\textsc{\textbf{ko}}&	\textsc{\textbf{fa}}&	\textsc{\textbf{Avg.}}\\
\midrule
XLM-R&	95.82 &	94.19 &	97.53 &	95.53 &	97.40 &	96.05 &	95.93 &	96.07  \\
Ensem&	96.56 &	95.11 &	97.83 &	\textbf{96.48} &	97.57 &	96.54 &	96.15 &	96.61  \\
ACE&	\textbf{96.69} &	\textbf{95.80} &	\textbf{98.22} &	96.46 &	\textbf{98.01} &	\textbf{96.79} &	\textbf{96.75} &	\textbf{96.96}  \\
\bottomrule
\end{tabular}
}
\caption{A comparison of ACE models, XLM-R models and an ensemble of the XLM-R models on the \textbf{development set}. }
\label{tab:ace}
\end{table}

\subsection{How Important Can the Types of KB be?}
\label{sec:context_type}
We compare several types of KBs and contexts during our system building. 
\paragraph{Online Search Engine}
In the early stage, we tried to use the knowledge retrieved from Google Search, which can retrieve related knowledge from a large scale of webs and is believed to be a strong multilingual search engine. 

\paragraph{Three Context Types Retrieved from Wikipedia}
As we mentioned in Section \ref{sec:post-processing}, there are three context processing options, which are: 1) use the matched paragraph; 2) use the matched sentence; 3) use the matched sentence but remove the wiki anchors. We denote the three options as \textsc{Para}, \textsc{Sent} and \textsc{Sent$_{\text{-link}}$} respectively.
\paragraph{Entity Retrieval with Gold Entities} We use gold entities on the development set to see whether the model performance can be improved. This can be seen as the most ideal scenario for iterative retrieval. We denote this process as \textsc{Iter}$_G$ and use \textsc{Para} for the context type.

In Table \ref{tab:context_comparison}, we can observe that: 1) For the three context options, \textsc{Para} is the best option for \textsc{\textbf{en}}, \textsc{\textbf{es}}, \textsc{\textbf{nl}}, \textsc{\textbf{ru}}, \textsc{\textbf{tr}}, \textsc{\textbf{ko}}, \textsc{\textbf{fa}}, \textsc{\textbf{mix}} and \textsc{\textbf{multi}}. \textsc{Sent$_{\text{-link}}$} is the best option for \textsc{\textbf{hi}} and \textsc{\textbf{bn}}. For \textsc{\textbf{de}} and \textsc{\textbf{zh}}, \textsc{Sent} and \textsc{Sent$_{\text{-link}}$} are competitive. As a result, we choose \textsc{Sent} for the two languages since we believe the wiki anchors from the Wikipedia can help model performance; 2) Comparing with the baseline, the knowledge from Google Search can improve model performance. Based on the best context option of each track, the knowledge from Wikipedia is better than the online search engine; 3) For \textsc{Iter}$_G$, we can find that the context can further improve the performance over 8 out of 13 tracks. However, there are only significant improvements for \textsc{\textbf{hi}}, \textsc{\textbf{bn}} and \textsc{\textbf{mix}}. 

\paragraph{Iterative Entity Retrieval with Predicted Entities} Based on the results in Table \ref{tab:context_comparison}, we further analyze how the predicted entity mentions can improve the retrieval quality. We denote the iterative entity retrieval with predicted mentions as \textsc{Iter}$_P$. In the experiment, we set $T=2$.\footnote{Our preliminary experiments show that there is no significant improvement for $T=3$.} We extract the predicted mentions of the development sets from the models based on the best context option for each track. We conduct the experiments over \textsc{\textbf{hi}}, \textsc{\textbf{bn}} and \textsc{\textbf{mix}} which have significant improvement with \textsc{Iter}$_G$. In Table \ref{tab:context_coarse2fine}, we also list the performance of \textsc{Iter}$_G$ for reference, which can be seen as using the predicted mentions with 100\% accuracy. From the results, we observe that only \textsc{\textbf{mix}} can be improved. 

Since iterative entity retrieval uses predicted mentions as a part of retrieval query, the performance of mention detection directly affects the retrieval quality. To further analyze the observation in Table \ref{tab:context_coarse2fine}, we evaluate the mention F1 score of the NER models with sentence retrieval. For comparison with mention detection performance of NER models, we additionally train mention detection models by discarding the entity labels during training. From the results in Table \ref{tab:mention}, we suspect the low mention F1 introduces noises in the knowledge retrieval module for \textsc{\textbf{bn}} and \textsc{\textbf{hi}}, which lead to the decline of performance as shown in Table \ref{tab:context_coarse2fine}. Moreover, the mention F1 of mention detection models (second row of Table \ref{tab:mention}) only outperform that of the NER models (first row of Table \ref{tab:mention}) in a moderate scale. Therefore, we train the \textsc{Iter} models only for the code-mixed track and use the NER models with sentence retrieval to predict mentions.

\subsection{Model Efficiency}
\label{sec:speed}
Table \ref{tab:speed} shows the speed of each module in our system. In the table, we also show that the retrieval speed of our local KB is significantly faster than that of Google Search. The bottleneck of the system speed is the NER module rather than the knowledge retrieval module. The main reason for the slow speed of the NER module is that the input length of the knowledge-based system is significantly longer than the original input. Taking the \textsc{\textbf{en}} test set as an example, there are on average 10 tokens for each input sentence in the original test set while there are 218 tokens for the input of our knowledge-based system. The longer inputs slow down the encoding at XLM-R embeddings.

\subsection{Effect of Embedding Concatenation}
\label{sec:ce}
We compare with some variants of our system that we designed but did not use in the test phase.

\paragraph{CE (Concatenation of Embeddings)} CE is one of the usual approaches to NER, which concatenates different kinds of embeddings to improve the token representations. In the early stage of our system building, we compare CE with only using the XLM-R embeddings based on the knowledge retrieved from the Google Search. Results in Table \ref{tab:google_ce} show that CE models are stronger than the models using XLM-R embeddings only in all the cases, which show the effectiveness of CE. 

\paragraph{ACE (Automated Concatenation of Embeddings)} ACE \citep{wang2020automated} is an improved version of CE which automatically selects a better concatenation of the embeddings. We use the same embedding types as CE and the knowledge are from our Wikipedia KB. We experiment on \textsc{\textbf{en}}, \textsc{\textbf{es}}, \textsc{\textbf{nl}}, \textsc{\textbf{ru}}, \textsc{\textbf{tr}}, \textsc{\textbf{ko}} and \textsc{\textbf{fa}}, which are strong with \textsc{Para} contexts. In Table \ref{tab:ace}, we further compare ACE with ensemble XLM-R models. Results show ACE can improve the model performance and even outperform the ensemble models\footnote{Please refer to Appendix \ref{sec:set_ce} for detailed settings.}. 

The results in Table \ref{tab:google_ce} and \ref{tab:ace} show the advantage of the embedding concatenation. However, as we have shown in Section \ref{sec:speed}, the prediction speed is quite slow with the single XLM-R embeddings. The CE models further slow down the prediction speed since the models contain more embeddings. The ACE models usually have faster prediction speed than the CE models. However, training the ACE models is quite slow. It takes about four days to train a single ACE model. Moreover, the ACE models cannot use the development set to train the model since they use development score as the reward to select the embedding concatenations. Therefore, due to the time constraints, we did not use these two variants in our submission during the shared task period.


\subsection{Effectiveness of Multi-stage Fine-tuning}
In Table \ref{tab:continue_tuning}, we show the effectiveness of multi-stage fine-tuning on the development set for our baseline system. The result shows that multi-stage fine-tuning can significantly improve the model performance for all the tracks.


\section{Conclusion}
In this paper, we describe our knowledge-based system for the MultiCoNER shared task, which wins 10 out of 13 tracks in the shared task. We construct multilingual KBs and retrieve the related documents from KBs to enhance the token representations of input text. We show that the NER models can use the retrieved knowledge to facilitate complex entity prediction, significantly improving both the in-domain and out-of-domain performance. Multi-stage fine-tuning can help the monolingual models learn from the training data of all the languages and improve the model performance and training efficiency. We also show that the system presents a good balance between the model performance and prediction efficiency to meet the time requirement in the test phase. We believe this system can be widely applied to other domains for the task of NER. For future work, we plan to improve the retrieval quality and adopt the system to support other kinds of entity-related tasks.



\section*{Acknowledgements}
This work was supported by Alibaba Group through Alibaba Innovative Research Program.

\bibliography{anthology,custom,aaai22,acl2021}
\bibliographystyle{acl_natbib}

\appendix
\section{Detailed Experimental Setup}
\label{app:exp}
The detailed statistics of the MultiCoNER dataset are listed in Table \ref{tab:stat} and the statistics of our KBs ares shown in Table \ref{tab:wiki}.

\subsection{Statistics of Datasets and Knowledge Bases}
\begin{table}[h]
\centering
\small
\begin{tabular}{lrrr}
\toprule
Track & {{Train}} & {{Dev}} & Test\\
\midrule
English & 15,300 & 800 & 217,818\\
Spanish & 15,300 & 800 & 217,887\\
Dutch & 15,300 & 800 & 217,337\\
Russian & 15,300 & 800 & 217,501\\
Turkish & 15,300 & 800 & 136,935\\
Korean & 15,300 & 800 & 178,249\\
Farsi & 15,300 & 800 & 165,702\\
German & 15,300 & 800 & 217,824\\
Chinese & 15,300 & 800 & 151,661\\
Hindi & 15,300 & 800 & 141,565\\
Bangla & 15,300 & 800 & 133,119\\
Multilingual & 168,300 & 8,800 & 471,911\\
Code-mixed & 1,500 & 500 & 100,000\\
\bottomrule
\end{tabular}
\caption{Statistics of the the MultiCoNER dataset (\# of sentences). Note that the training and development sets of the multilingual dataset are a mixture of monolingual training and development sets respectively.}
\label{tab:stat}
\end{table}

\begin{table}[h]
\centering
\small
\begin{tabular}{lrrrr}
\toprule
Language & Pages & Paragraphs & ES Docs\\
\midrule
English & 8,075,229 & 138,259,937 & 224,077,884 \\
Spanish & 1,813,109 & 29,767,543 & 47,248,391 \\
Dutch & 2,234,442 & 18,007,520 & 29,442,016 \\
Russian & 2,437,595 & 44,536,255 & 77,903,362 \\
Turkish & 728,950 & 8,196,825 & 12,685,674 \\
Korean & 905,976 & 11,965,418 & 16,326,787 \\
Farsi & 1,502,301 & 13,723,218 & 17,342,825 \\
German & 3,147,933 & 54,315,261 & 98,386,199 \\
Chinese & 1,659,253 & 20,342,685 & 14,888,964 \\
Hindi & 196,745 & 1,926,636 & 3,279,827 \\
Bangla & 203,869 & 2,526,333 & 4,342,959 \\
\bottomrule
\end{tabular}
\caption{Detailed statistics on 11 languages.}
\label{tab:wiki}
\end{table}

\subsection{System Configurations}
For the knowledge retrieval module, we retrieve top-10 related results from the KB. For iterative entity retrieval, we set $T=2$.
In masked language model pretraining, we use a learning rate of $5\times 10^{-5}$.
For the NER module, we use a learning rate of $5\times 10^{-6}$ for fine-tuning the XLM-R embeddings and use a learning rate of $0.05$ to update the parameters in the CRF layer following \citet{wang-etal-2021-improving}. Each NER model built by our system can be trained and evaluated on a single Tesla V100 GPU with 16GB memory. For the ensemble module, we train about 10 models for each track. 

\subsection{Settings of CE and ACE models}
\label{sec:set_ce}
In Section \ref{sec:ce}, we compare our NER model with CE and ACE models. In CE and ACE models, we concatenate monolingual fastText \citep{bojanowski2017enriching} word embeddings, monolingual/multilingual Flair embeddings \citep{akbik-etal-2018-contextual}, ELMo embeddings \citep{peters-etal-2018-deep,che-EtAl:2018:K18-2}, XLM-R embeddings fine-tuned on the whole training data and XLM-R embeddings fine-tuned on the language data by multi-stage fine-tuning. We only feed the knowledge-based input into XLM-R embeddings and feed the original input into other embeddings because it is hard for the other embeddings (especially for LSTM-based embeddings such as Flair and ELMo) to encode such a long input. We use Bi-LSTM encoder to encode the concatenated embeddings with a hidden state of 1,000 and then feed the output token representations into the CRF layer. Following most of the previous efforts, we use SGD optimizer with a learning rate of 0.01. For ACE, we search the embedding concatenation for 30 episodes.

\section{More Analysis}

\label{app:more_anal}

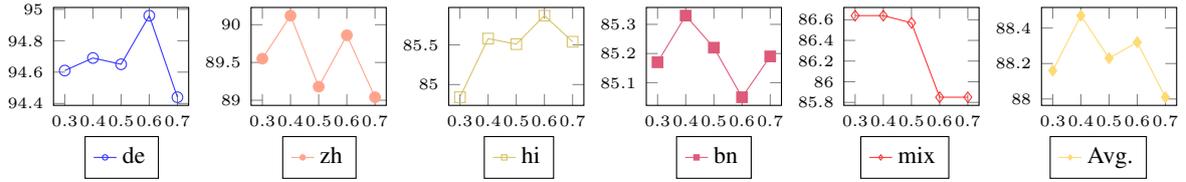
\begin{figure*}[t!]
\begin{minipage}{1.0\linewidth}
\centering
\begin{tikzpicture}
    \begin{axis}[
        xshift=-10cm,
        name=ner,
        width=0.21\textwidth,
        height=0.18\textwidth,
        legend columns=2, 
        legend image post style={scale=0.5},
        legend style={font=\small,at={(0.78,-0.75)}, anchor=south east},
        tick label style={font=\tiny},
        xtick={0.3,0.4,0.5,0.6,0.7},
        ylabel style={font=\tiny,yshift=-0.2cm},
        ]
        \addplot[blue!80,mark=o] table[x=threshold,y=de] {twitter15.dat};
        \legend{de}
    \end{axis}
    \begin{axis}[
        at={(ner.south west)},
        xshift=2.6cm,
        width=0.21\textwidth,
        height=0.18\textwidth,
        legend columns=2, 
        legend image post style={scale=0.5},
        legend style={font=\small,at={(0.78,-0.75)}, anchor=south east},
        tick label style={font=\tiny},
        xtick={0.3,0.4,0.5,0.6,0.7},
        ylabel style={font=\tiny,yshift=-0.5cm},
        ]
        \addplot[red!80!yellow!45,mark=*] table[x=threshold,y=zh] {twitter15.dat};
        \legend{zh}
    \end{axis}
    \begin{axis}[
        at={(ner.south west)},
        xshift=5.2cm,
        width=0.21\textwidth,
        height=0.18\textwidth,
        legend columns=2, 
        legend image post style={scale=0.5},
        legend style={font=\small,at={(0.78,-0.75)}, anchor=south east},
        tick label style={font=\tiny},
        xtick={0.3,0.4,0.5,0.6,0.7},
        ylabel style={font=\tiny,yshift=-0.5cm},
        ]
        \addplot[yellow!80!blue!90,mark=square] table[x=threshold,y=hi] {twitter15.dat};
        \legend{hi}
    \end{axis}
    \begin{axis}[
        at={(ner.south west)},
        xshift=7.8cm,
        width=0.21\textwidth,
        height=0.18\textwidth,
        legend columns=2, 
        legend image post style={scale=0.5},
        legend style={font=\small,at={(0.78,-0.75)}, anchor=south east},
        tick label style={font=\tiny},
        xtick={0.3,0.4,0.5,0.6,0.7},
        ylabel style={font=\tiny,yshift=-0.5cm},
        ]
        \addplot[red!80!blue!65,mark=square*] table[x=threshold,y=bn] {twitter15.dat};
        \legend{bn}
    \end{axis}
    \begin{axis}[
        at={(ner.south west)},
        xshift=10.4cm,
        width=0.21\textwidth,
        height=0.18\textwidth,
        legend columns=2, 
        legend image post style={scale=0.5},
        legend style={font=\small,at={(0.78,-0.75)}, anchor=south east},
        tick label style={font=\tiny},
        xtick={0.3,0.4,0.5,0.6,0.7},
        ylabel style={font=\tiny,yshift=-0.5cm},
        ]
        \addplot[red!80,mark=diamond] table[x=threshold,y=mix] {twitter15.dat};
        \legend{mix}
    \end{axis}
    \begin{axis}[
        at={(ner.south west)},
        xshift=13.0cm,
        width=0.21\textwidth,
        height=0.18\textwidth,
        legend columns=2, 
        legend image post style={scale=0.5},
        legend style={font=\small,at={(0.78,-0.75)}, anchor=south east},
        tick label style={font=\tiny},
        xtick={0.3,0.4,0.5,0.6,0.7},
        ylabel style={font=\tiny,yshift=-0.5cm},
        ]
        \addplot[yellow!80!red!65,mark=diamond*] table[x=threshold,y=avg] {twitter15.dat};
        \legend{Avg.}
    \end{axis}
\end{tikzpicture}
\caption{An illustration of majority voting threshold versus the ensemble model performance. }
\label{fig:ensemble_curve}
\end{minipage}
\end{figure*}

\begin{table}[t!]
\small
\setlength\tabcolsep{4pt}
\centering
\begin{tabular}{lccccc|c}
\toprule
&	\textsc{\textbf{de}}&	\textsc{\textbf{zh}}&	\textsc{\textbf{hi}}&	\textsc{\textbf{bn}}&	\textsc{\textbf{mix}}&	\textsc{\textbf{Avg.}}  \\
\midrule
Voting&	\textbf{94.65} &	\textbf{89.18} &	\textbf{85.51} &	\textbf{85.22} &	\textbf{86.57} &	\textbf{88.23}  \\
CRF&	94.04 &	88.96 &	85.37 &	85.12 &	85.33 &	87.76  \\
\bottomrule
\end{tabular}
\caption{A comparison of ensemble approaches on the \textbf{development set}. }
\label{tab:ensemble}
\end{table}

\subsection{Majority Voting Ensemble and CRF Level Ensemble}
As we state in Section \ref{sec:ensemble}, we use majority voting as the ensemble algorithm in our system. We show an experiment about how the voting threshold affect the ensemble model performance during our system building on the development set. We ensemble the models on \textsc{\textbf{de}}, \textsc{\textbf{zh}}, \textsc{\textbf{hi}}, \textsc{\textbf{bn}}, \textsc{\textbf{mix}} with \textsc{Para} since these five tracks have relatively lower performance than the other 7 tracks. In Figure \ref{fig:ensemble_curve}, we show how the threshold of the majority voting affects the model performance. From the figure, we can see that the best threshold varies over the language. Therefore, we simply choose 0.5 as there is no best threshold value. Moreover, we compare the majority voting ensemble and CRF level ensemble in Table \ref{tab:ensemble}. The CRF level ensemble averages the emission and transition scores in the Eq. \ref{eq:psi} predicted by the candidate models and uses the Viterbi algorithm to get the prediction. The results show that CRF level ensemble performs inferior to the majority voting ensemble. The possible reason is that training with different random seeds may lead to different emission transition scores at different scales. As a result, the models with larger scales have higher weights in the ensemble.

\begin{table}[t!]
\small
\setlength\tabcolsep{2pt}
\centering
\begin{tabular}{lcccc}
\toprule
Test Context  &	\multicolumn{2}{c}{\textsc{Para}}&	\multicolumn{2}{c}{Opt$_\text{Best}$} \\
Search KB&		All&	Language&	All&	Language \\
\midrule
Wiki-\textsc{Para}&	84.57&	84.94&	-&	- \\
Wiki+Opt$_\text{Best}$&	-&	84.96&	84.38&	84.78 \\
\bottomrule
\end{tabular}
\caption{Test results for multilingual models with different context options and different KB size. }
\label{tab:multilingual}
\end{table}

\subsection{How the Search Space and the Context Type Affects Multilingual Model Performance?}
In the multilingual test set, we can find 304,905 sentences in the other monolingual test sets while there are 167,006 sentences that cannot be found. For these sentences, we can either search on the whole KB of all languages or first detect the language of the input sentence and then search in the specific language KB\footnote{We determine the language of the input sentence using the langdetector (\url{https://pypi.org/project/langdetect/}) tool.}. Moreover, as we discussed in Section \ref{sec:context_type}, using different kinds of retrieved knowledge affects the model performance. As a result, we train two types of multilingual models. One is only using the \textsc{Para} contexts for all language and another is using the best option for each language based on Table \ref{tab:context_comparison}. From the results in Table \ref{tab:multilingual}, we can observe that: 1) searching over the language specific KB performs better than searching the whole KB, 2) using the language specific context option cannot improve the model performance. Therefore, we ensemble both types of the model for the final submission. 
\end{document}

%% file: math_commands.tex
\usepackage{amsmath,amsfonts,bm}

\def\eqref#1{equation~\ref{#1}}

\def\1{\bm{1}}

\def\vv{{\bm{v}}}

\def\vx{{\bm{x}}}
\def\vy{{\bm{y}}}

\DeclareMathAlphabet{\mathsfit}{\encodingdefault}{\sfdefault}{m}{sl}
\SetMathAlphabet{\mathsfit}{bold}{\encodingdefault}{\sfdefault}{bx}{n}

\def\sR{{\mathbb{R}}}

